%% file: main.tex
\documentclass[letterpaper, 10 pt, conference]{ieeeconf}  

\pdfoutput=1

\IEEEoverridecommandlockouts                              

\usepackage{times} 

\usepackage[ruled,linesnumbered]{algorithm2e}
\usepackage{algpseudocode}

\usepackage{amsmath} 
\usepackage{amssymb}  
\usepackage{color}
\usepackage{graphicx}
\usepackage{subfigure}
\usepackage[english]{babel}
\usepackage{amsthm}
\usepackage{breqn}
\usepackage[super]{nth}
\usepackage{url}
\usepackage{siunitx}
\usepackage{makecell}
\usepackage{microtype}
\usepackage{booktabs}
\usepackage{hyperref}
\usepackage{stfloats}
\usepackage{threeparttable} 
\usepackage{multirow}
\theoremstyle{definition}

\usepackage{verbatim}

\usepackage[utf8]{inputenc}
\usepackage{graphicx}
\usepackage{subfigure}
\usepackage{float}
\usepackage{bm}
\usepackage{amssymb}
\usepackage{threeparttable}
\usepackage{multirow}
\usepackage{eqnarray}
\usepackage{color}
\usepackage{graphicx}
\usepackage{tablefootnote}

\usepackage{balance}
\makeatletter
\newcommand{\removelatexerror}{\let\@latex@error\@gobble}
\makeatother

\title{\bf Flow-Aware Navigation of Magnetic Micro-Robots in Complex Fluids via PINN-Based Prediction}
\author{Yongyi Jia, Shu Miao, Jiayu Wu, Ming Yang, Chengzhi Hu, and Xiang Li
\thanks{
Y. Jia, S. Miao, J. Wu, and X. Li are with the Department of Automation, Tsinghua University. M. Yang and C. Hu are with Department of Mechanical and Energy Engineering, College of Engineering, Southern University of Science and Technology. This work was supported in part by the Science and Technology Innovation 2030-Key Project under Grant 2021ZD0201404, in part by
Beijing National Research Center for Information Science and Technology, in part by the National Natural Science Foundation of China under Grant U21A20517 and 62461160307, and in part by the BNRist project under Grant BNR2024TD03003.
Corresponding author: Xiang Li (xiangli@tsinghua.edu.cn)
}}
\begin{document}

\maketitle
\thispagestyle{empty} 
\pagestyle{empty}  

\newtheorem{definition}{Definition}

\begin{abstract}
While magnetic micro-robots have demonstrated significant potential across various applications, including drug delivery and microsurgery, the open issue of precise navigation and control in complex fluid environments is crucial for in vivo implementation. This paper introduces a novel flow-aware navigation and control strategy for magnetic micro-robots that explicitly accounts for the impact of fluid flow on their movement. First, the proposed method employs a Physics-Informed U-Net (PI-UNet) to refine the numerically predicted fluid velocity using local observations. Then, the predicted velocity is incorporated in a flow-aware A* path planning algorithm, ensuring efficient navigation while mitigating flow-induced disturbances. Finally, a control scheme is developed to compensate for the predicted fluid velocity, thereby optimizing the micro-robot's performance. A series of simulation studies and real-world experiments are conducted to validate the efficacy of the proposed approach. This method enhances both planning accuracy and control precision, expanding the potential applications of magnetic micro-robots in fluid-affected environments typical of many medical scenarios.
\end{abstract}

\section{Introduction}
\input{chapters/1_introduction}
\section{Related Works}
\input{chapters/2_related_works}
\section{Methods}

\input{chapters/3_METHODOLOGY}

\section{Results}
\input{chapters/4_results}

\balance
\section{Conclusions}
\input{chapters/5_conclusions}
 
\clearpage

{\small
\bibliographystyle{IEEEtran}
\bibliography{ref}
}
\end{document}

%% file: chapters/1_introduction.tex
Magnetic micro-robots have attracted significant attention due to their vast potential in a variety of micromanipulation applications, particularly in the medical field, where they are poised to revolutionize procedures such as targeted drug delivery \cite{feng2024weak}, microsurgery \cite{acemoglu2017magnetic}, and cell transport \cite{miao2021development}. Typically controlled by external magnetic fields, these robots are capable of maneuvering through narrow and complex environments \cite{yu2021adaptive}, offering the promise of precise, non-contact, and damage-free transport, especially for in vivo applications. Nevertheless, the realization of fully automated navigation and control within intricate fluidic environments, such as blood vessels or interstitial spaces, remains challenging, primarily due to the complex interaction between the micro-robots and the surrounding fluid dynamics.

Existing methods for magnetic micro-robot navigation often assume static fluid conditions \cite{fang2024autonomous, yang2022autonomous}, in which the fluid's effect on the robot’s movement is treated as a simple disturbance or external force. For example, some existing approaches ignore flow influence and rely on disturbance-observer-based controllers \cite{zhong2023spatial} that treat fluid velocity as a perturbation, compensating for flow-induced effects. These methods are prone to flow reversal and struggle to handle higher flow rates. In other approaches, simulations incorporating flow fields are employed, and reinforcement learning (RL) algorithms are used to adapt robot behavior to dynamic fluid environments \cite{abbasi2024autonomous, colabrese2017flow}. However, these approaches require extensive training in simulated environments, and the gap between simulated and real-world conditions complicates deployment. End-to-end reinforcement learning methods have been proposed to address autonomous navigation in complex vascular pathways \cite{yang2022hierarchical, amoudruz2024path}. Under carefully designed reward functions, the robot can learn the optimal path while navigating safely within appropriate flow velocities and implicitly handling dynamic obstacles. However, these methods suffer from poor generalization across different vascular structures, hindering real-world deployment.

\begin{figure}[!t]
    \centering   \hspace{-5pt} \includegraphics[width=8cm]{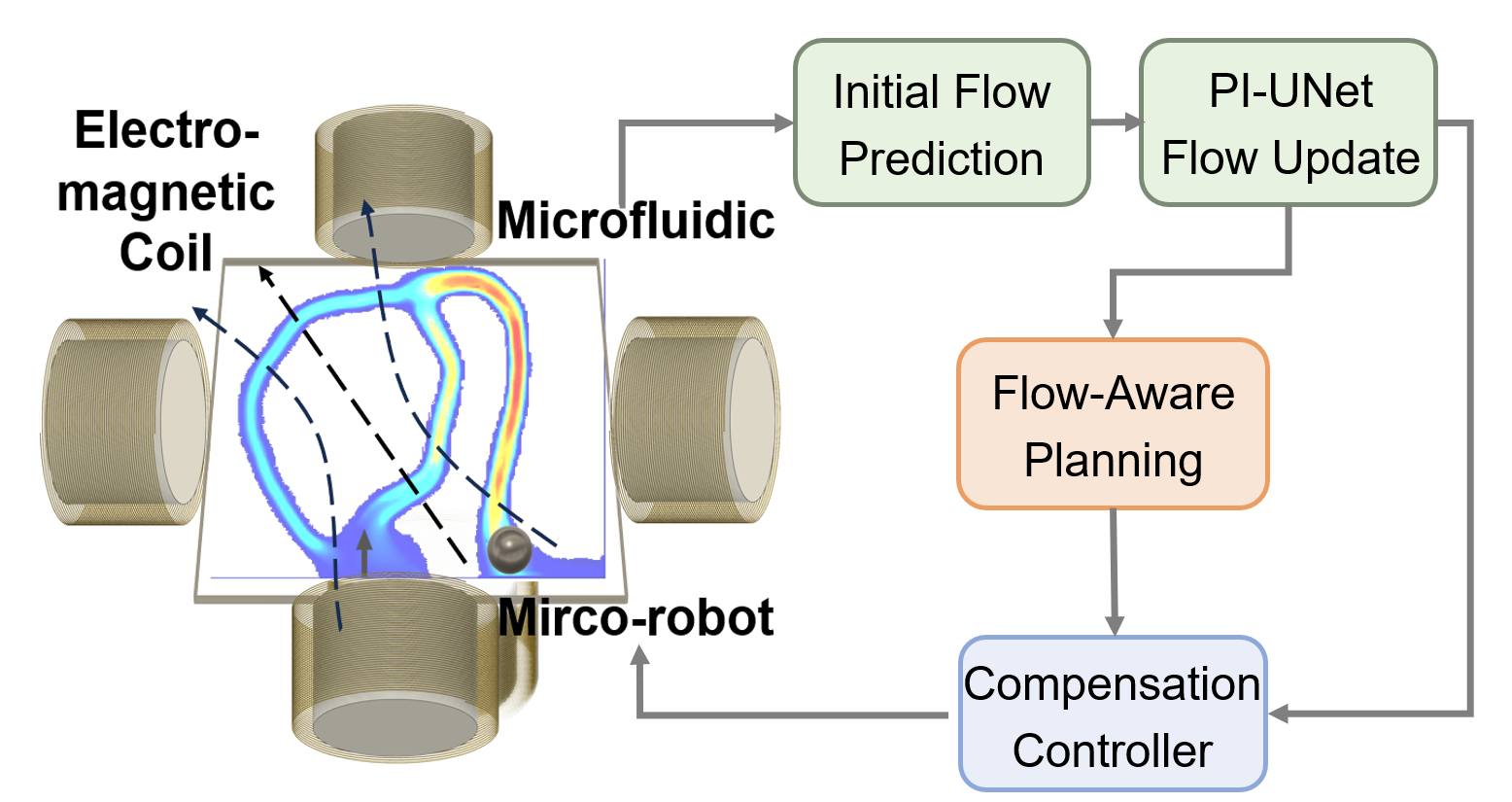}

    \caption{Overview of the proposed framework. The flow velocity is initially predicted and refined using the Finite Volume Method (FVM) and Physics-Informed Neural Networks (PINNs), then leveraged for path planning and disturbance compensation in dynamic microenvironments.}

    \label{fig:frame}
\end{figure}

To address these limitations, this paper introduces a novel flow-aware navigation and control method for magnetic micro-robots that directly incorporates fluid velocity into the decision-making process.
Unlike previous methods that either oversimplify the effects of fluid flow or rely on reinforcement learning with limited real-world applicability, our approach provides a more accurate and adaptable solution that accounts for the dynamic nature of fluid flows. 
The contributions of this paper are summarized as follows:
\begin{itemize}
\item[-] Flow Prediction: a novel approach combining the Finite Volume Method (FVM) and Physics-Informed Neural Networks (PINNs) is introduced to predict and refine fluid velocity in dynamic complex environments.
\item[-] Path Planning: the predicted flow is integrated into an A* algorithm to enable efficient and accurate path planning in fluid-filled environments.
\item[-] Velocity Compensation Controller: a control strategy is developed to counteract fluid-induced disturbances, ensuring robust and precise micro-robot navigation.
\end{itemize}

The overall framework of the proposed method is illustrated in Fig. \ref{fig:frame}. To validate the proposed approach, simulations and experimental results are presented, demonstrating its effectiveness in realistic scenarios such as navigation through blood vessel-like environments. This work introduces a comprehensive framework for flow-aware navigation and control, enabling more precise and adaptable applications in fluid-dominated settings, and lays the foundation for more generalized control systems in complex, real-world applications.

%% file: chapters/2_related_works.tex
\subsection{Navigation and Control of Magnetic Micro-robots in Fluid Environments}

Autonomous navigation in complex environments represents a crucial milestone in advancing micro-robotics toward intelligent applications. Various macroscopic path planning algorithms have been adapted for microscopic applications. For example, search-based methods such as the A* algorithm \cite{fan2018automated}, sample-based approaches like Rapidly-exploring Random Trees (RRT) \cite{huang2017path}, and zero-order optimization techniques exemplified by Particle Swarm Optimization (PSO) \cite{wang2021micromanipulation} have all been investigated for the navigation of micro-robots. Simultaneously, several local planning strategies have been developed to enhance obstacle avoidance and multi-agent coordination, including artificial potential field methods \cite{lee2021real}, trajectory optimization techniques \cite{li2018development}, and radar-based approaches \cite{3263773}. Additionally, reinforcement learning has been introduced to enable micro-robots to navigate complex vascular channels \cite{abbasi2024autonomous} and cluttered environments \cite{yang2022hierarchical}. Furthermore, it has been applied to directly map observations into low-level control policies \cite{10161023}. Although certain methods incorporate fluid dynamics within simulation environments \cite{amoudruz2024path}, most of these navigation strategies assume a static fluid setting, which leads to significant discrepancies when transitioning to real-world and in vivo applications.

After planning the reference trajectory, the direction and strength of the magnetic field should be designed to control the robot's trajectory tracking. The robot's kinematics is typically modeled as a two-dimensional, underactuated, nonholonomic system, where the flow field and other noise are treated as unknown disturbances. Trajectory tracking is usually formulated as a discrete point reach problem, where the simplest controller design method involves a fixed speed and the optimization of motion direction based on distance and directional errors \cite{9740251}. Additionally, other nonlinear control methods, such as feedforward-feedback control \cite{zhong2023spatial}, adaptive sliding mode control \cite{qi2024robust}, and virtual reference feedback tuning \cite{wang2024data}, are widely used to improve path tracking performance in complex environments or with unknown model parameters. In these control methods, disturbance observers are often designed to monitor and explicitly compensate for environmental disturbances. However, introducing higher-order terms can increase system instability, especially in the presence of high-frequency or nonlinear disturbances.

\subsection{Learning Based Flow Velocity Prediction}

The flow velocity in human blood vessels and similar channels satisfies the Navier-Stokes equation, and its simulation or prediction is typically treated as a partial differential equation (PDE) problem. Traditional methods for solving this problem are based on numerical simulations, such as finite difference methods \cite{causon2010introductory}, finite element methods \cite{brooks1982streamline}, and boundary element methods \cite{liu2011recent}. Although classical numerical methods achieve high precision, incorporating noisy real-world data into the algorithm remains challenging, hindering effective data assimilation \cite{sharma2023review}. Physics-informed neural networks (PINNs) \cite{raissi2019deep}, can seamlessly integrate multimodality experimental data with the various Navier-Stokes formulations for incompressible flows \cite{cai2021physics}. PINNs model the mapping from spatial coordinates to physical quantities using fully connected neural networks, in which the residuals of the Navier-Stokes equations and boundary conditions are incorporated into the loss function. The network parameters are updated through gradient descent to solve the associated PDE problem. Thus, observational data can be integrated into the network as either soft or hard constraints. Convolutional neural networks and generative models have been developed to improve the scalability and convergence of PINNs, such as PhyCRNet \cite{ren2022phycrnet}, PhyGeoNet \cite{gao2021phygeonet} and DiffusionPDE \cite{huang2024diffusionpdegenerativepdesolvingpartial}.

Recent studies have used PINNs for the prediction of flow velocity in hemodynamic applications. For example, weighted extended and conservative PINNs have been developed for detailed arterial blood flow simulations using generalized space-time domain decomposition \cite{bhargava2024enhancing}. Transfer learning frameworks have also been introduced to accelerate real-time blood flow predictions in ultra-fast ultrasound imaging \cite{guan2023towards}. In addition, PINNs have been used to predict arterial blood pressure from 4D flow MRI data by enforcing conservation laws \cite{KISSAS2020112623}, and to infer brain hemodynamics from sparse measurement data \cite{9740143}. Furthermore, a point cloud-based PINNs approach has been explored for the prediction of 4D hemodynamics, highlighting the importance of vascular morphology \cite{zhang2023physics}. Despite these advancements, current PINNs methods are predominantly applied to problems with simple boundaries, such as the two-dimensional flow around a cylinder, and remain challenged by the complexities of multibranch flow channels.

%% file: chapters/3_METHODOLOGY.tex

\subsection{PI-UNet Based Flow Prediction}
\label{unet}

Micro-robotic systems typically operate in low-flow-velocity microenvironments with low Reynolds number \cite{xu2013modeling}. Therefore, this study considers a two-dimensional steady laminar flow within vascular environments, which satisfies the Navier-Stokes equations. The flow velocity prediction problem is formulated as the estimation of the two-dimensional flow velocity maps, $\mathcal{V}_x$ and $\mathcal{V}_y$ in the $x$ and $y$ directions, respectively, given the boundary mask image $\mathcal{I}$ of the channel and local observations $\{\bm{y}_{obs}: \bm{v}_{obs}\}$. The solution is constrained to satisfy the corresponding governing equations,
\begin{equation}
    \frac{\partial v_x}{\partial x} + \frac{\partial v_y}{\partial y} = 0
    \label{eq:continuty}
\end{equation}
\begin{equation}
    \rho \left( v_x\frac{\partial v_x}{\partial x} + v_y\frac{\partial v_x}{\partial y} \right) = - \frac{\partial p}{\partial x} + \nu \left(\frac{\partial^2 v_x}{\partial^2 x} + \frac{\partial^2 v_x}{\partial^2 y}\right)
    \label{eq:x}
\end{equation}
\begin{equation}
    \rho \left( v_x\frac{\partial v_y}{\partial x} + v_y\frac{\partial v_y}{\partial y} \right) = - \frac{\partial p}{\partial y} + \nu \left(\frac{\partial^2 v_y}{\partial^2 x} + \frac{\partial^2 v_y}{\partial^2 y}\right)
    \label{eq:y}
\end{equation}
where (\ref{eq:continuty}) represents the continuity equation, (\ref{eq:x}) and (\ref{eq:y}) are the momentum equations in the $x$ and $y$ directions, respectively, $v_x$, $v_y$ are the continuous velocity components in the $x$ and $y$ directions, $p$ is the pressure, $\rho$ is the fluid density, and $\nu$ is the kinematic viscosity. 

\begin{figure}[!t]
    \centering    \hspace{-1pt}\includegraphics[width=8.7cm]{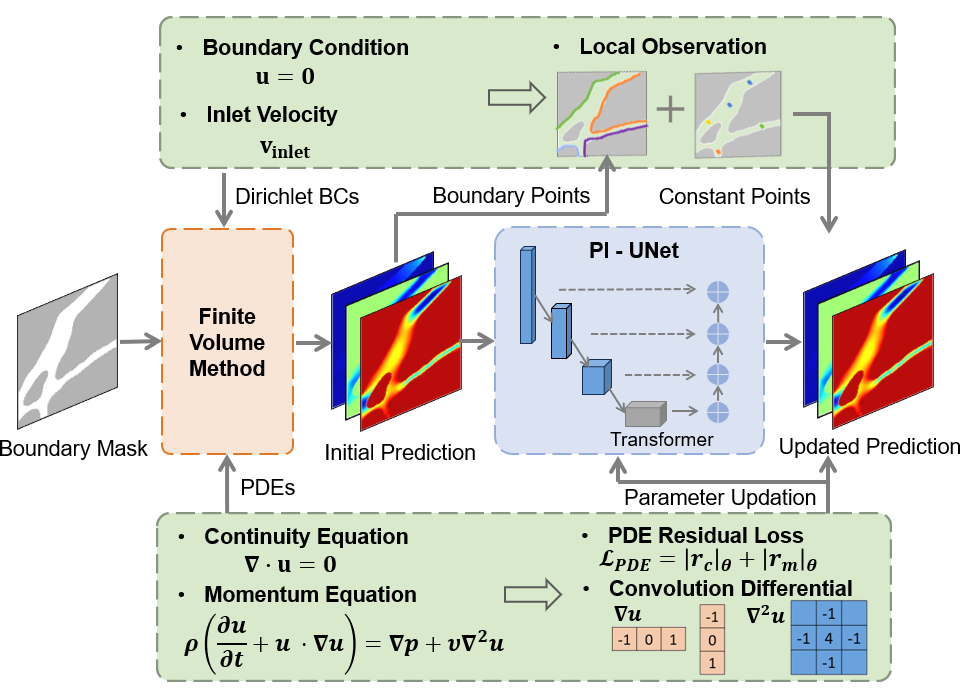}
    \caption{The flow prediction framework. The velocity prediction is formulated as a PDE problem. The initial velocity map is obtained using the finite volume method (FVM) and then refined by the PI-UNet model, which integrates local observations and PDE residual loss for improved accuracy.}
    \label{fig:model}
\end{figure}

First, the PDE problem is solved by the Finite Volume Method (FVM) under the assumed fluid density $\hat \rho$ and viscosity $\hat \nu$, and the initial estimation of velocity is obtained. Specifically, the boundary image $\mathcal{I}$ is divided into small control surfaces. The governing equations are integrated over each control surface $A$. For the continuity equation, it becomes
\begin{equation}
    \int_A \left( \frac{\partial v_x}{\partial x} + \frac{\partial v_y}{\partial y} \right) dA = 0.
\end{equation}
By applying Gauss's divergence theorem, the area integral is converted into a line integral along the one-dimensional boundary $S$ of the control volume
\begin{equation}
    \oint_S \left( v_x \hat{n}_x + v_y \hat{n}_y \right) dS = 0,
\end{equation}
where $\hat{n}_x, \hat{n}_y$ are the components of the outward unit normal vector on the boundary, and $dS$ is the differential length element along the boundary. For the momentum equations, the flux terms in the momentum equations are discretized using similar area integrals. The above process transforms the PDEs into a set of linear algebraic equations for the discretized physical quantities $v_x$, $v_y$, and $p$ on each control face. Combined with the non-slip condition (\ref{eq: non-slip}) on the boundary and the estimated inlet velocity $v_{\text{inlet}}$, the linear algebraic equations can be solved iteratively. 
\begin{equation}
    v_x=0, \quad v_y=0 \quad \text{on} \quad \partial \mathcal{I} 
    \label{eq: non-slip}
\end{equation}
After obtaining the physical quantity of each grid point, the linear interpolation generates the initial velocity prediction maps $\mathcal{V}_{x, \text{ini}}$, $\mathcal{V}_{y, \text{ini}}$ and $\mathcal{P}_{\text{ini}}$.

Due to factors such as fluid properties and boundary conditions, the fluid velocity in real-world environments may deviate from the results obtained through finite element analysis. In this context, local flow velocity in the actual environment can be observed through robot state feedback or Doppler ultrasound velocity measurement \cite{wang2021flexible}, leading to partially observable challenges. For instance, the next section presents a method for observing flow velocity based on robot state. To achieve more accurate estimation and effectively utilize observational data, we design a flow velocity update network based on the Physics-Informed UNet (PI-UNet) to optimize the flow velocity.

To better encode complex channel geometries, traditional continuous PINNs are replaced with a convolutional neural network based on discrete images. The primary structure of the network combines a four-layer UNet architecture with a Transformer module. As shown in Fig. \ref{fig:model}, the network takes in the boundary mask and the initial prediction as inputs shaped $[H, W, 4]$, producing an optimized prediction $\mathcal{V}_{x, \text{ opt}}$, $\mathcal{V}_{y, \text{opt}}$ and $\mathcal{P}_{\text{opt}}$ as the output. To preserve global prediction performance and prevent an excessive focus on local information, the Transformer mechanism is incorporated only at the U-Net's bottleneck layer rather than at every layer as in Attention-UNet \cite{oktay2018attention}. Skip connections are formed by concatenating features from the encoder to the corresponding decoder layers, allowing the network to retain fine-grained spatial details.

The loss function is composed of the residuals of the governing equation (\ref{eq:continuty}), (\ref{eq:x}), and (\ref{eq:y}), where the first-order and second-order differential functions are computed by gradient-free convolution filters, and the convolution kernels are given by
\begin{equation}
    K_{\partial} = \frac{1}{2}
    \begin{bmatrix}
       -1 & 0 & 1     
    \end{bmatrix} ,
    \label{eq: first-order}
\end{equation}
\begin{equation}
    K_{\partial^2} = \frac{1}{4}
    \begin{bmatrix}
        0 & -1 & 0          \\
       -1 & 4 & -1     \\
        0 & -1 & 0
    \end{bmatrix} .
    \label{eq: second-order}
\end{equation}
We construct a steady optimization problem by strictly enforcing the observational data onto the network output, ensuring that the observational data can influence the global pixel node predictions through the gradient backpropagation of the loss function. Similarly, the initially predicted boundary pixel values are strictly integrated into the network output as Dirichlet boundary conditions for the PDE problem. Studies indicate that providing values near the boundary, rather than merely imposing zero-velocity boundary conditions at the boundary points, facilitates the solution of the differential equation within complex geometric channels.

\subsection{Flow-Aware Navigation and Control for Micro-robots}

\begin{figure}[!t]
    \centering    \includegraphics[width=8.5cm]{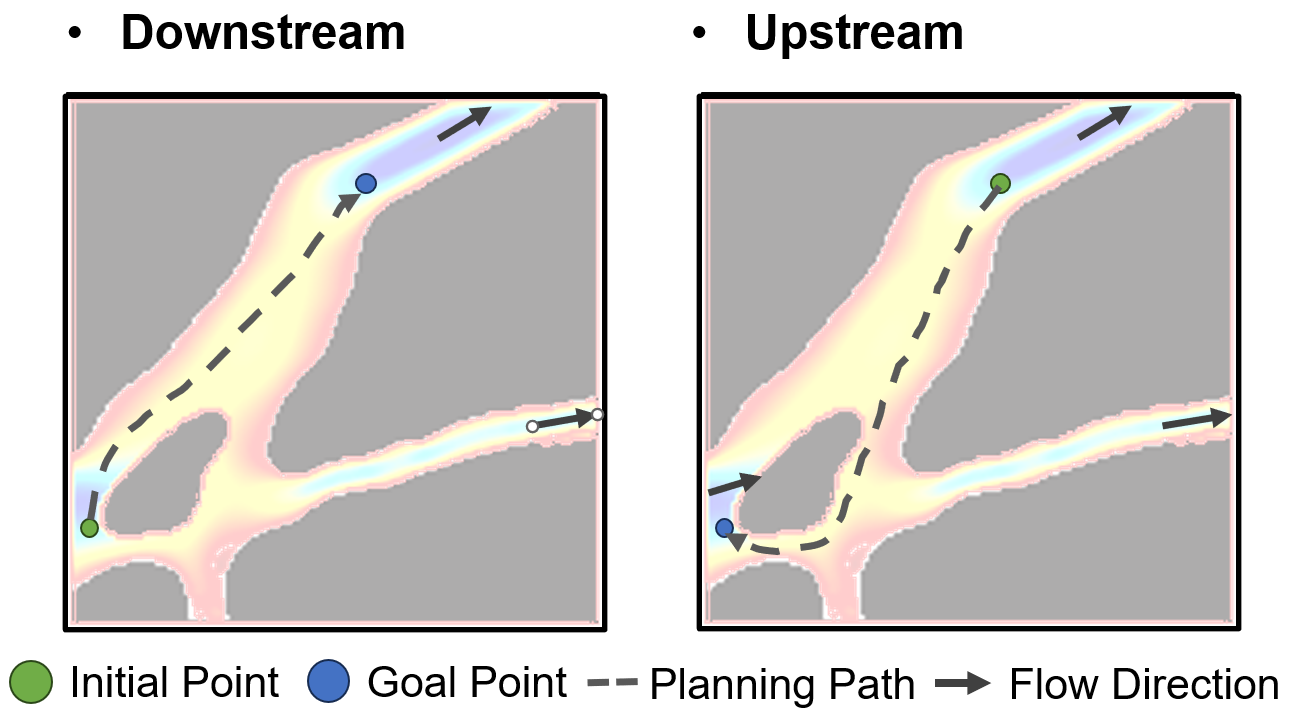}
    \caption{Example of robot navigation in the flow channel. The background color represents flow velocity, where different paths are chosen in downstream and upstream scenarios to minimize travel time.}
    \label{fig:flow}
\end{figure}

In this section, we explore how flow prediction can support the design of navigation and control algorithms for micro-robot. Specifically, we consider a genearl simplified two-dimensional micro-robot system driven by a rotating magnetic field with a kinematic model of
\begin{equation} 
\bm{\dot x} = 
    \begin{bmatrix}
       {\dot x}    \\
       {\dot y}
    \end{bmatrix} =
    \begin{bmatrix}
       \cos(\phi) u +v_x   \\
     \sin(\phi) u + v_y
    \end{bmatrix}, 
     \label{eq: kinematic}
\end{equation}
where $u$ represents the robot's forward velocity, which is directly proportional to the rotation frequency of the magnetic field $f$; $\phi$ is the yaw angle of the magnetic field; and $v_x$ and $v_y$ represent the components of the fluid flow velocity defined in the previous subsection. The kinematic model is highly versatile and is applicable to various forms of micro-robots, including spherical, helical, and swarm-based micro-robotic systems. Figure \ref{fig:flow} presents a simplified example to illustrate the impact of flow direction on the micro-robot's path selection. In the downstream scenario (left), the robot is supposed to choose a shorter route with higher flow velocity to reduce travel time. Conversely, in the upstream scenario (right), a longer path with lower flow velocity should be selected to ensure a safer and more reliable arrival.

To achieve the above outcome, we propose a flow-aware navigation algorithm to plan an optimal path that minimizes the robot's travel time or energy consumption. Given the narrow and geometrically complex nature of the flow channels, sampling-based motion planning methods are not suitable. Additionally, the highly nonlinear nature of the flow velocity presents challenges for the application of the artificial potential field method. Therefore, we develop a flow-aware A* algorithm in a discretized space, incorporating flow velocity into both the heuristic and cost functions to optimize path planning. To control the number of search points, we use grid points generated by FVM discretization, as the search nodes. Each node and its corresponding flow rate are represented as a pair $\{\bm{x}_{\text{node}}: \bm{v}_{\text{node}}\}$. A K-D Tree is employed to construct the adjacency graph, enabling efficient nearest-neighbor searches. The A* algorithm works by expanding nodes from the initial node with the least total estimated cost $f(n)$, which is the sum of the actual cost $g(n)$ from the start node to the current node and the heuristic cost $h(n)$ to the goal node $n_{\text{goal}}$
\begin{equation}
f(n)=g(n)+h(n).    
\end{equation}
The actual cost of moving from a current node to a neighboring node is calculated by 
\begin{equation}
c(n_{\text{cur}}, n_{\text{neig}}) = \|\bm{x}_{\text{neig}} - \bm{x}_{\text{cur}}\|\cdot\|\frac{\bm{x}_{\text{neig}} - \bm{x}_{\text{cur}}}{\|\bm{x}_{\text{neig}} - \bm{x}_{\text{cur}}\|} - \frac{\bm v_{\text{cur}}}{v_{\text{max}}}\|,
\end{equation}
where $v_{\text{max}}$ is the maximum velocity of the micro-robot. The first term represents the Euclidean distance, and the second term quantifies the difference between the unit direction vector and the unit flow velocity vector, determining the required magnetic field compensation for reaching the target. Similarly, the heuristic cost is defined as
\begin{equation}
    h(n_{\text{cur}}) = \|\bm{x}_{\text{goal}} - \bm{x}_{\text{cur}}\|\cdot\|\frac{\bm{x}_{\text{goal}} - \bm{x}_{\text{cur}}}{\|\bm{x}_{\text{goal}} - \bm{x}_{\text{cur}}\|} - \frac{\bm v_{\text{cur}}}{v_{\text{max}}}\|.
\end{equation}
The algorithm explores the path by expanding the node with the smallest $f(n)$ value until the goal node is found or all nodes are explored. At each step, A* adds the current node's neighbors to the open list, updates their cost, and moves the processed nodes to the closed list, ensuring the search for the optimal path.

After obtaining the reference trajectory $\bm{x}_d$, a feedforward-feedback controller with flow compensation is designed based on the kinematic model (\ref{eq: kinematic}). Unlike a single-input controller which only considers the yaw angle for tracking discrete points, our controller simultaneously computes both the input velocity and the yaw angle, enabling trajectory tracking over time. The feedforward velocity $u_{\text{ff}}$ and heading $\phi_{\text{ff}}$ are computed based on the difference between the desired velocity and the flow velocity as
\begin{equation}
    u_{\text{ff}} = \|\dot{\bm{x}}_d - \bm{v}\|,
\end{equation}
\begin{equation}
    \phi_{\text{ff}} = \arctan(\dot{\bm{x}}_d - \bm{v}).
\end{equation}
The position error in the global frame is computed as $\bm{e} = \bm{x}_d - \bm{x} - \bm{v}\cdot dt$ with $dt$ denoting the time interval, and subsequently transformed into a local coordinate system defined by the reference yaw angle $\phi_d = \arctan2(\bm{\dot x}_d)$, as
\begin{equation}
    \bm{e}^{\text{loc}} = \begin{bmatrix}
        {e}^{\text{loc}}_x\\{e}^{\text{loc}}_y
    \end{bmatrix} = 
    \begin{bmatrix}
        \cos \phi_d & \sin \phi_d \\
        -\sin \phi_d & \cos \phi_d
    \end{bmatrix} \begin{bmatrix}
        e_x\\
        e_y
       \end{bmatrix} .
\end{equation}
The feedback control for the heading $\phi_{\text{fb}}$ is obtained as
\begin{equation}
    \phi_{\text{fb}} = \arctan(\bm{e}).
\end{equation}
Then, the yaw angle input $\phi$ is computed as a weighted average of the feedforward and feedback headings as 
\begin{equation}
    \phi = \frac{|e^{\text{loc}}_x|}{|e^{\text{loc}}_x|+|e^{\text{loc}}_y|}\phi_{\text{ff}} +\frac{|e^{\text{loc}}_y|}{|e^{\text{loc}}_x|+|e^{\text{loc}}_y|}\phi_{\text{fb}},
    \label{eq: phi}
\end{equation}
where the coefficients calculated by the local error are used to balance the tracking of the reference trajectory and the reduction of the absolute error. The feedback term for velocity $u_{\text{bf}}$ is designed to reduce the error in the forward direction as
\begin{equation}
    u_{\text{fb}} = K_p\left(e_x\cos\phi+e_y\sin\phi\right),
\end{equation}
where $K_p$ is the proportional coefficient. The total velocity input is obtained by the combination of the feedforward and feedback terms
\begin{equation}
    u = u_{\text{ff}} + u_{\text{fb}}.
\end{equation}
Finally, based on the kinematic model (\ref{eq: kinematic}), a straightforward disturbance observer is derived from the kinematic model (\ref{eq: kinematic}) to estimate the local flow velocity as 
\begin{equation}
    \dot{\hat {\bm{v}} }= \bm{L}_p (\dot {\bm{x}} - \dot{\hat{\bm{x}}}),
    \label{eq: est}
\end{equation}
where $\bm{L}_p$ is the coefficient matrix and $\hat{\bm{x}}$ is the estimated state. The resulting estimate can then be used to refine the flow velocity prediction discussed in subsection~\ref{unet}.

%% file: chapters/4_results.tex
\subsection{Simulation Studies}
\begin{figure*}[!t]
    \centering
    \includegraphics[width=16cm]{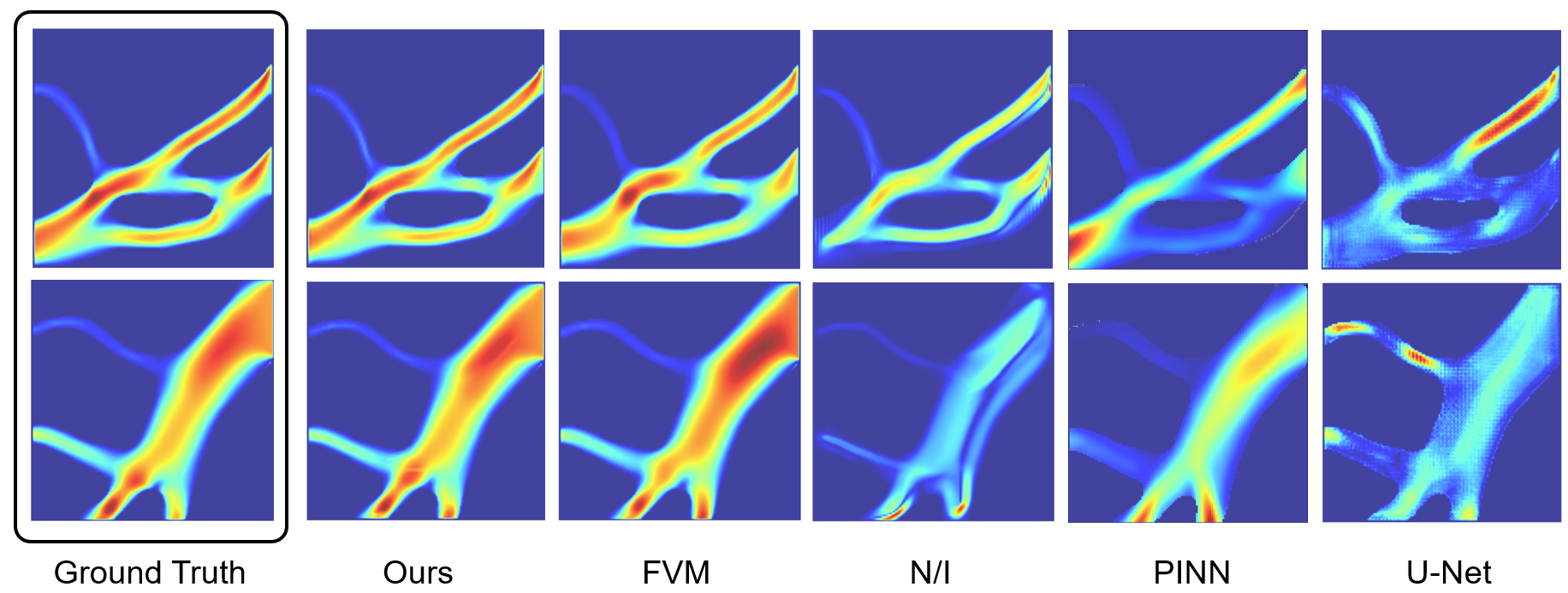}
    \vspace{-5pt}
    \caption{Comparison of different flow prediction methods across two cases. The first column represents the ground truth. Flow velocity is normalized within the range of 0 to 1, with colors transitioning from blue (low velocity) to red (high velocity).}
    \label{fig: pred}
\end{figure*}

We first construct a flow channel dataset and evaluate different velocity prediction methods. The dataset is obtained by randomly cropping images from the entire fundus vasculature \cite{amoudruz2024path}, with each sample of size $[128,128]$. To simulate a partially observable real-world fluid environment, we generate realistic velocity fields by randomly modifying the fluid properties in COMSOL \cite{comsol2024} to generate realistic velocity fields. Some observational data are extracted from the centerline of a pixel-wide flow channel. We compare four baseline methods: 1) FVM: utilizes only the initial prediction from FVM; 2) N/I: no initial prediction, relying only on observed data with PI-UNet; 3) PINN: a PINN model based on continuous coordinates; 4) U-Net: a data-driven U-Net model without physics constraints. As shown in Fig. \ref{fig: pred}, our proposed method is in closest agreement with the velocity distribution of the ground truth. The conventional FVM exhibits deviations in high-velocity regions (red zones) due to the absence of local measurements; N/I introduces artificial discontinuities as it fails to extrapolate flow fields from boundary observations; PINN struggles to enforce physical constraints on complex geometric boundaries, performing well only in data-rich regions, such as near inlets and along boundaries; U-Net captures velocity gradients but fails to reconstruct the true distribution, while also introducing grid-like artifacts. Table \ref{tab: error} presents the mean and standard deviation of the relative root mean square error (RMSE) across 10 images. Our method achieves the lowest error in $v_x$, $v_y$, and the total velocity $\bm{v}$, outperforming all baselines. These results demonstrate its effectiveness in integrating local observations with physical constraints while handling complex geometric flow channels. 

\begin{table}[ht]
\caption{Comparison of RMSE Between Our Method and Baselines}

    \begin{center} 
    \resizebox{0.47 \textwidth}{!}{
    \begin{threeparttable}
    \centering
        \begin{tabular}{ccccc}
        \toprule[1pt]
        \multicolumn{2}{c}{\multirow{2}{*}{RMSE $\downarrow$}}   & \multicolumn{3}{c}{Flow Velocity}   \\ 
         &     & {$v_x$}      & {$v_y$}       &{$\bm{v}$}    \\ 
        \midrule[0.6pt]
        \multirow{5}{*}{Method}      
        & Ours  & $\bm{0.101 \pm 0.033} $ & $\bm{0.102\pm0.056}$ & $\bm{0.102 \pm 0.044}$ \\ 
        & FVM  & $ 0.162 \pm 0.071$  & $0.191 \pm 0.122$   & $0.173 \pm 0.096$   \\
        & N/I  & $ 0.673 \pm 0.214$  & $0.795 \pm 0.172$   & $0.713 \pm 0.197$   \\
        & PINN  & $ 0.541 \pm 0.067$  & $0.575 \pm 0.069$   & $0.556 \pm 0.062$   \\
        & U-Net  & $ 0.865 \pm 0.032$  & $0.800 \pm 0.012$   & $0.665 \pm 0.011$   \\
       \bottomrule[1pt] 
       \end{tabular}
    \end{threeparttable}}
    \end{center} 
    \label{tab: error}

\end{table}

\begin{figure}[tb]
    \centering
    \includegraphics[width=8.4cm]{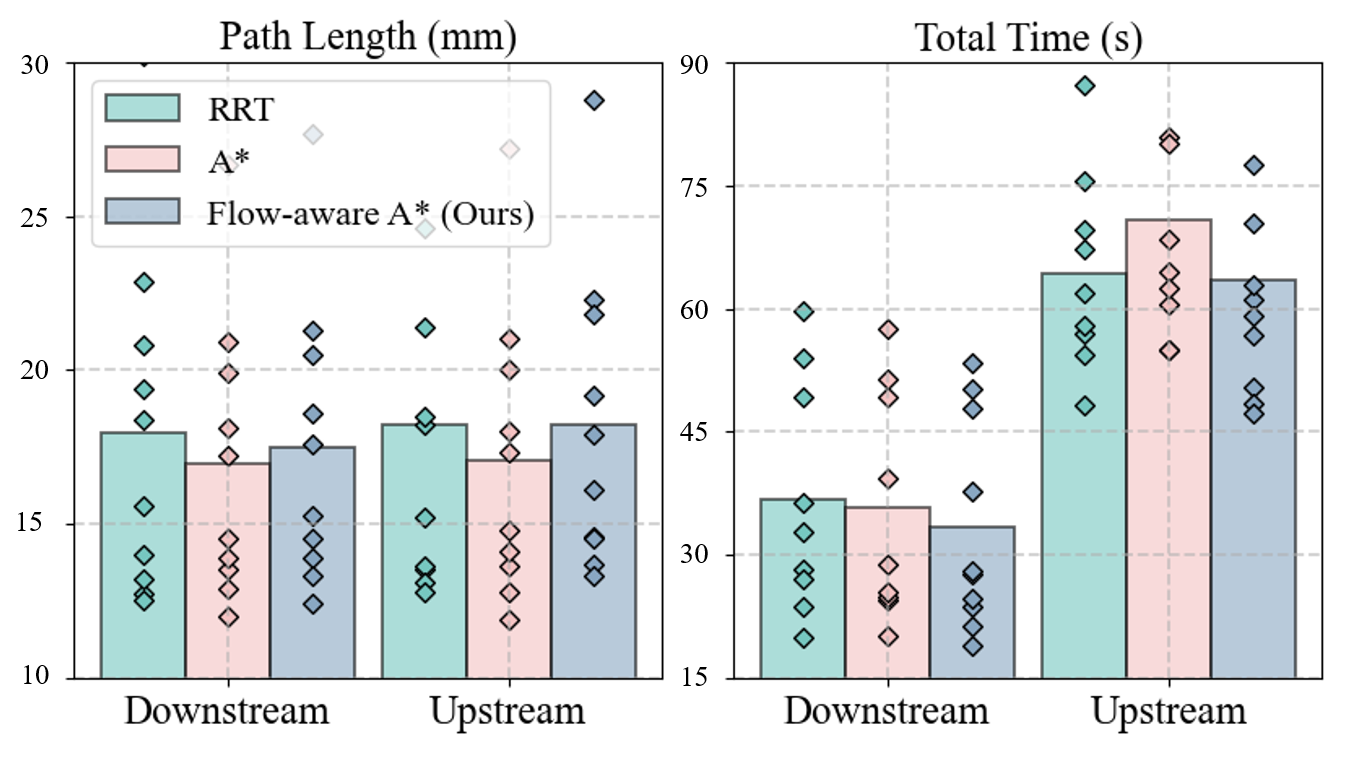}
    \caption{Evaluation of the path planning with different methods and scenarios. Each bar shows the average path length / time over 10 channels in the dataset, and the values for each channel are also plotted as scattered diamond-shaped points.}
    \label{fig: plan}
\end{figure}

\begin{figure}[tb]
    \vspace{-3pt}
    \centering
    \includegraphics[width=8.3cm]{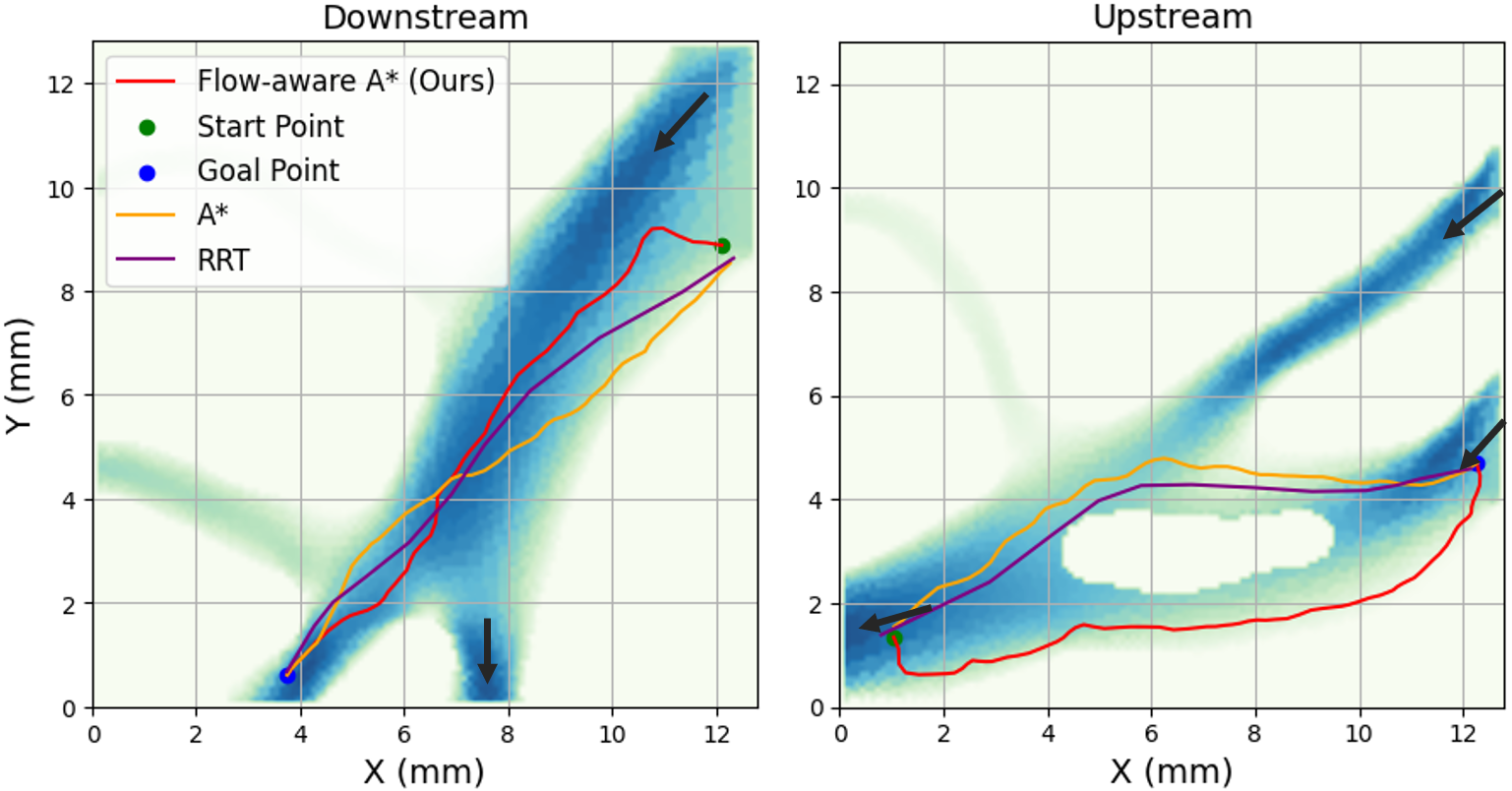}
    \caption{Visualization of planned paths using different methods. Solid lines represent the smoothed paths, the background indicates flow velocity magnitude, and black arrows denote the flow direction.}
    \vspace{-3pt}
    \label{fig: path}
\end{figure}

The flow-aware path planning method based on the above-predicted velocity dataset is then implemented and compared with the A* and RRT methods. In each image, two distant points are selected as start and goal positions to simulate both downstream and upstream navigation scenarios. Figure \ref{fig: plan} presents the planned path lengths and the corresponding travel times under identical input conditions. Since A* considers only Euclidean distance, it generates the shortest paths. The time required for upstream navigation is approximately twice that of downstream, highlighting the influence of the flow field on the robot. Our method, which explicitly accounts for flow velocity, achieves the shortest travel time. Figure \ref{fig: path} illustrates two representative scenarios. In the downstream case, our planner directs the robot toward the high-velocity core before following the flow downstream. Conversely, in the upstream scenario, the planner selects lower-velocity pathways and navigates along the boundaries to minimize the impact of the flow field.

\begin{figure}[tb]
    \centering
    \hspace{-2pt}
    \includegraphics[width=8.4cm]{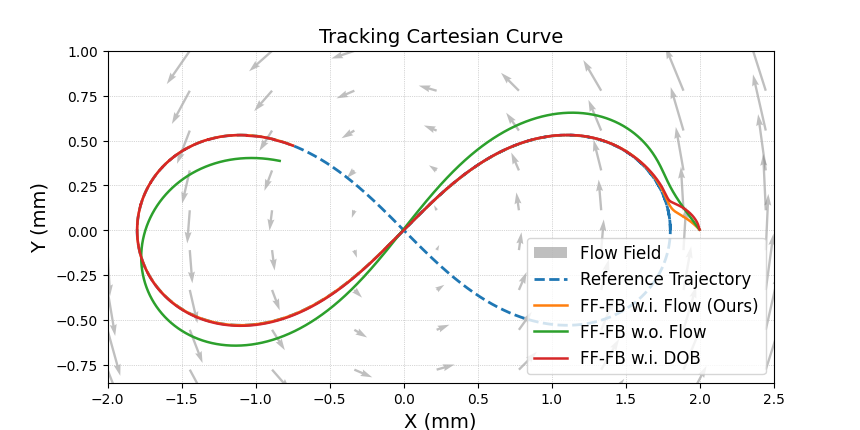}
    \caption{Trajectory tracking path. The gray arrow shows the spatial distribution of the flow field. All three controllers are initialized at the same position, which is offset from the reference trajectory.}
    \label{fig: traj}
\end{figure}

\begin{figure}[tb]
    \centering
    \includegraphics[width=8cm]{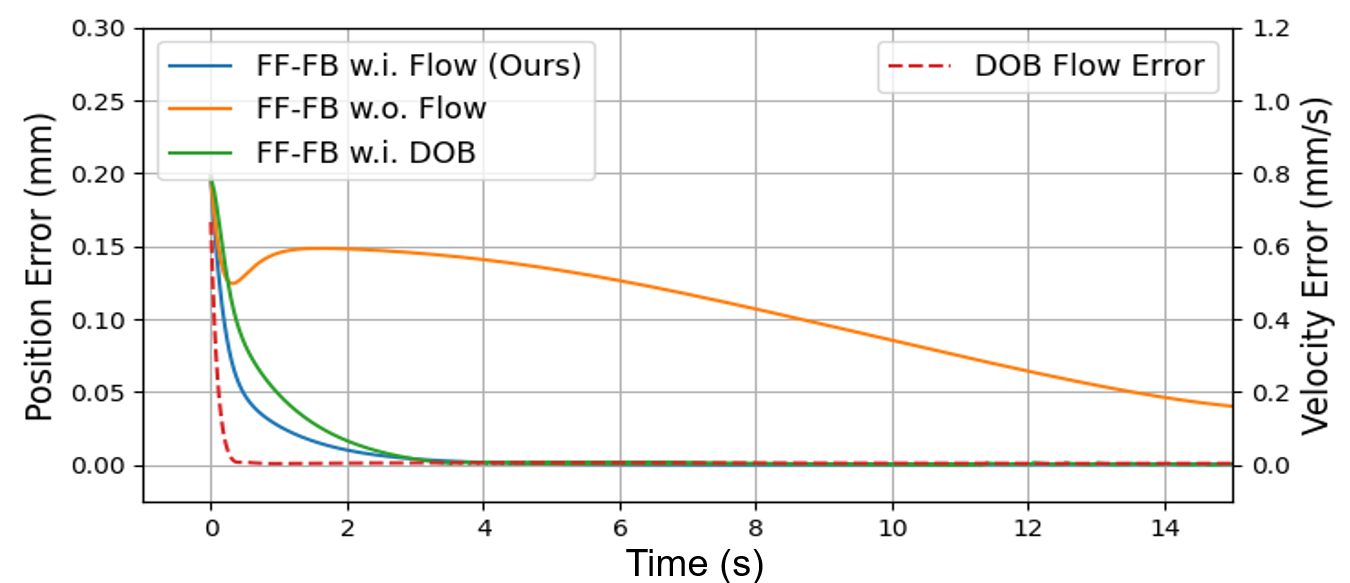}
    \caption{Tracking error curve over time. The solid line represents the trajectory error, while the dashed line represents the flow velocity error of the disturbance observer.}
    \label{fig: error}
\end{figure}

The performance of the trajectory tracking controller is evaluated through numerical simulations. Two baseline controllers are realized: 1) feedforward-feedback controller without flow compensation; 2) feedforward-feedback controller with disturbance observer described in (\ref{eq: est}). The controller hyperparameters are set to ${K}_p =1$ and $\bm{L}_p = \text{diag}(1, 1)$. The spatially correlated flow field is assumed to be $\bm{v} = [-0.2y, 0.2x]$ and the reference trajectory is set as 
\begin{equation}
    \bm{x}_d = 
    \begin{bmatrix}
    \frac{a \cos(\theta)}{1 + \sin(\theta) \sin(\theta)}  \\[10pt]
    \frac{b \sin(\theta) \cos(\theta)}{1 + \sin(\theta) \sin(\theta)}
    \end{bmatrix} \quad \hspace{-0.3cm}\theta \in \left[0, 2\pi\right], a = 1.8 mm, b = 1.5 mm.
    \label{eq: ref_traj}
\end{equation}
As illustrated in Fig. \ref{fig: traj} and Fig. \ref{fig: error}, our FF-FB controller with flow compensation achieves rapid convergence of the tracking error. Under the same feedback gain, ignoring flow velocity leads to a persistent error. The disturbance observer provides accurate local flow velocity estimates, but using them directly in compensation slows convergence.
\begin{figure}[tb]
    \centering
    \includegraphics[width=8cm]{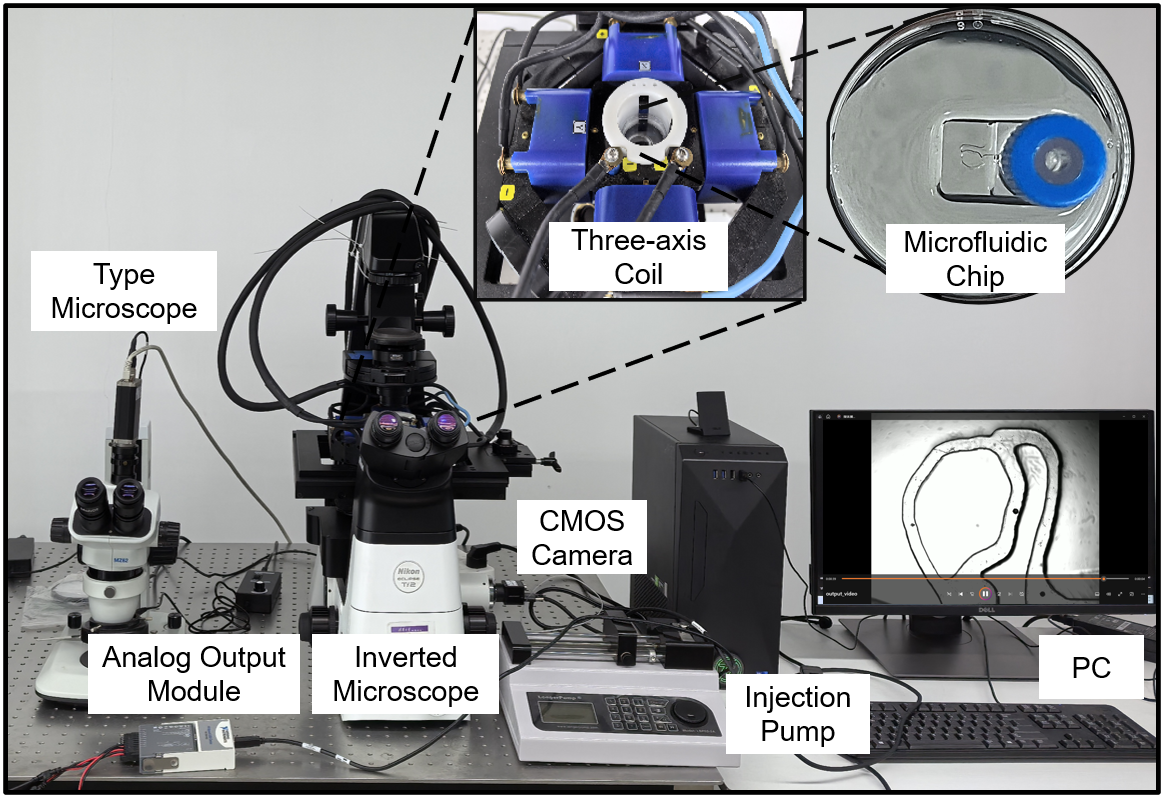}
    \vspace{-5pt}
    \caption{The hardware setup used in the experiments consists of an inverted microscope equipped with a three-axis coil and an analog output module, a PC with an integrated camera, and a microfluidic chip connected to an injection pump.}

    \label{fig: setup}
\end{figure}

\subsection{Real-World Experiments}

\begin{figure*}[tb]
    \centering
    \includegraphics[width=16cm]{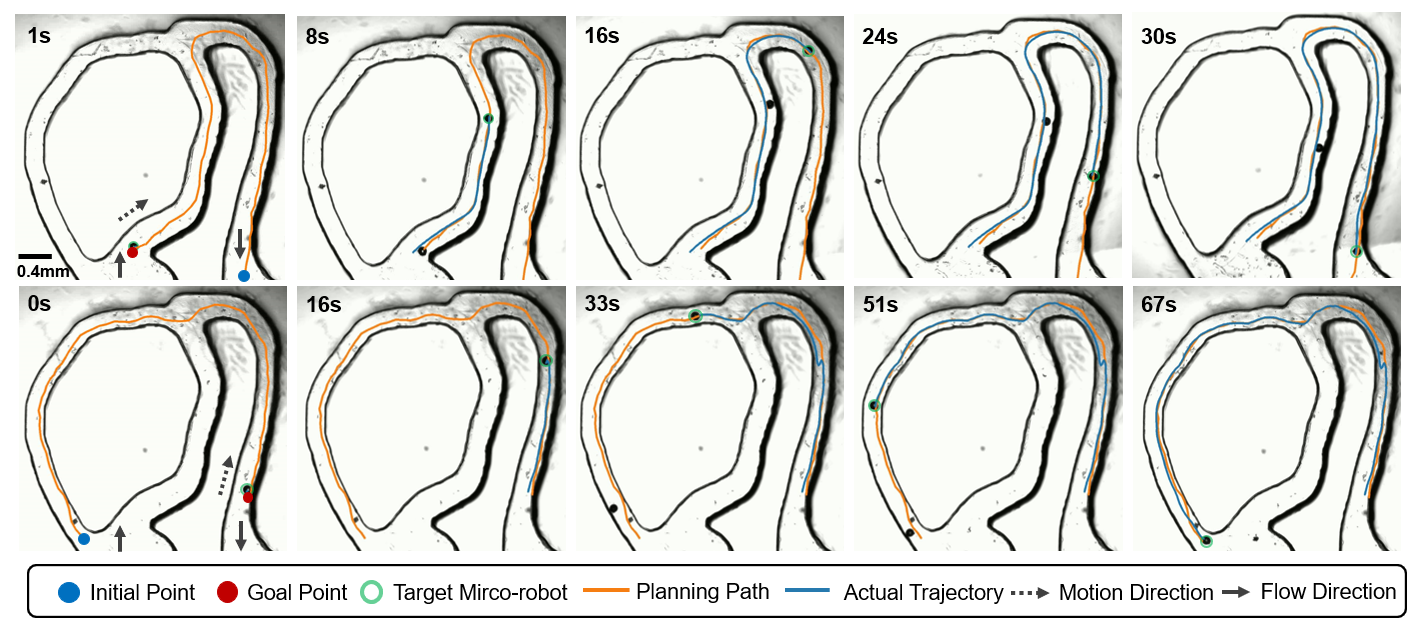}
    \caption{Experiment of spherical magnetic driven robot navigation in the microchannel. The first column corresponds to the initial frame, where the robot’s position is identified through image processing and the goal position is manually assigned. The top row illustrates downstream navigation from the chip inlet to the outlet, while the bottom row shows upstream navigation from the outlet back to the inlet.}
    \label{fig: test}
\end{figure*}

We conduct experiments on a magnetically driven micro-robot system, as shown in Fig. \ref{fig: setup}. The hardware setup consists of a three-axis electromagnetic coil setup mounted on an inverted microscope (Nikon, ECLIPSE-Ti2). A CMOS camera (HIKROBOT, MV-CH050-10UM) captures microscopic images and transmits them to a PC (Intel i7-9750H CPU), where control inputs are computed. These signals are then output via an analog module (NI-9263) to generate the rotating magnetic field. A PMMA microfluidic chip is manufactured, having the narrowest channel of $0.2 mm$, and a mechanical syringe pump (LONGER, LSP01-3A) is attached to the inlet to simulate the flow environment. The micro-robots are prepared by sputtering a $200 nm$ nickel coating onto $100 \mu m$ silica microspheres and manually isolated under a type microscope.

\begin{figure}[tbh]

    \centering
    \includegraphics[width=8cm]{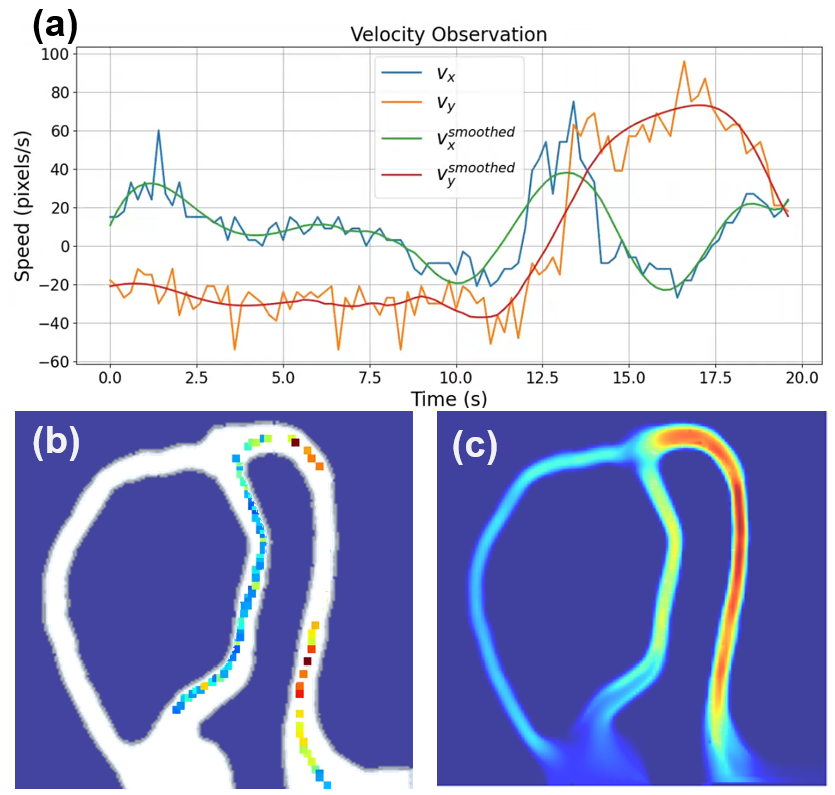}
    \caption{Flow velocity estimation and refinement. (a) Fluid velocity variation over time. (b) Flow magnitudes along the robot’s trajectory, serving as local observation data. (c) Refined flow velocity predictions generated by PI-UNet.}

    \label{fig: flow_update}
\end{figure}

We collect the velocity observations of the robot within the channel to estimate the flow speed. The robot travels downstream under the combined influence of a manually configured magnetic field and the surrounding flow. Figure \ref{fig: flow_update} (a) depicts fluid velocity variation over time. To reduce noise, the top and bottom $20\%$ of the velocity values are discarded, followed by polynomial smoothing. Figure \ref{fig: flow_update} (b) presents the flow magnitudes recorded along the robot’s trajectory, serving as local observation data. Finally, Figure \ref{fig: flow_update} (c) shows the refined flow velocity predictions generated by PI-UNet.

Based on the predicted flow velocity, path planning is performed, and the magnetic field is controlled to navigate the micro-robot to the target position. Figure \ref{fig: test} presents two sets of experimental results. In the first scenario, shown in the top row, the micro-robot moves downstream from the channel inlet to the outlet. At the bifurcation, the planner selects the shorter, higher-velocity path, enabling the robot to reach the target in 31 seconds. In the second scenario, depicted in the bottom row, the micro-robot navigates upstream from the outlet to the inlet. The flow-aware planner chooses the longer, lower-velocity left-side path to reduce flow resistance. Additionally, the planned trajectory follows the channel boundary to further mitigate the impact of the flow. As a result, the micro-robot completes the upstream navigation task in 67 seconds. More videos can be found in the supplementary materials.

%% file: chapters/5_conclusions.tex
This paper presents a flow-aware navigation and control framework for magnetic micro-robots, directly incorporating fluid dynamics to address disturbances in dynamic environments. By combining the Finite Volume Method (FVM) and physics-informed neural networks (PINNs), our approach achieves high-fidelity fluid flow prediction, enabling adaptive path planning via an A*-based algorithm designed to handle flow disturbances. A velocity compensation control scheme further mitigates deviations caused by fluid forces. Simulations and experiments in blood vessel-like environments validate the framework’s superiority over methods that oversimplify flow effects or depend on impractical learning strategies. The integration of flow prediction, efficient planning, and robust control improves navigation precision in complex fluid environments, supporting biomedical applications such as targeted drug delivery and cell surgery. This work establishes a foundational framework for autonomous navigation in fluid-dominated scenarios and has potential applications in in-body navigation of magnetic micro-robots.